\documentclass[11pt,a4paper]{article}
\usepackage[hyperref]{acl21}
\usepackage{times}
\usepackage{latexsym}
\usepackage{amsmath}
\usepackage{courier}
\usepackage{graphicx}
\usepackage{mathrsfs}
\usepackage{amsfonts}
\usepackage{subfigure}
\usepackage{multirow}
\usepackage{url} 
\usepackage{algorithm}
\usepackage{algorithmic}
\newcommand{\paratitle}[1]{\noindent\textbf{#1}}

\aclfinalcopy 

\author{
Wangchunshu Zhou\thanks{\ \ Equal contribution.} ~\thanks{\ \ Corresponding author} ~~~ Qifei Li$^{*}$ ~~~ Chenle Li\\
Beihang University, Beijing, China\\
{\tt zhouwangchunshu@buaa.edu.cn}}
\date{}

\begin{document}

\title{Learning from Perturbations: Diverse and Informative \\ Dialogue Generation with Inverse Adversarial Training}

\maketitle

\begin{abstract}
In this paper, we propose \textbf{I}nverse \textbf{A}dversarial \textbf{T}raining (IAT) algorithm for training neural dialogue systems to avoid generic responses and model dialogue history better. In contrast to standard adversarial training algorithms, IAT encourages the model to be sensitive to the perturbation in the dialogue history and therefore learning from perturbations. By giving higher rewards for responses whose output probability reduces more significantly when dialogue history is perturbed, the model is encouraged to generate more diverse and consistent responses. By penalizing the model when generating the same response given perturbed dialogue history, the model is forced to better capture dialogue history and generate more informative responses. Experimental results on two benchmark datasets show that our approach can better model dialogue history and generate more diverse and consistent responses. In addition, we point out a problem of the widely used maximum mutual information (MMI) based methods for improving the diversity of dialogue response generation models and demonstrate it empirically.
\end{abstract}
\section{Introduction}

In recent years, neural end-to-end dialogue response generation models~\cite{sordoni2015neural,serban2016building,bordes2016learning} has gained increasing popularity with the recent advancements of neural sequence-to-sequence (seq2seq) learning models~\cite{sutskever2014sequence,vaswani2017attention}. While neural dialogue models can generate seemingly fluent responses, due to the over-simplified maximum likelihood estimation (MLE) training objective and the high frequency of generic responses in training corpora, they tend to produce dull and generic responses such as ``I don't know'' much more often than that humans generally do~\cite{li2015diversity}, which makes dialogue agents less engaging and ineffective. 

In addition, recent research on whether neural dialogue systems use dialogue history effectively~\cite{sankar2019neural} shows that most neural dialogue agents fail to take the dialogue history into account when generating responses. This problem makes neural dialogue systems tend to generate responses irrelevant to the current topic of the conversation and are not consistent with the dialogue history. This problem may also intensify the generic response problem, as dull responses are generally off-topic and irrelevant to the dialogue history. 


To address the above issues, in this paper, we propose \textbf{I}nverse \textbf{A}dversarial \textbf{T}raining (IAT) algorithm for training neural dialogue systems to avoid generic responses and model dialogue history better, thus generating diverse and informative responses. Conventional adversarial training methods generally generate label-preserving adversarial inputs with carefully designed methods and train the model to generate the same output to enhance the model's robustness. In contrast, our approach perturbs in input dialogue history such that a good dialogue model should \textbf{not} generate the same output if the output is non-generic and relevant to the dialogue history. We name our proposed method as inverse adversarial training because it is related to conventional adversarial training methods which aim to improve the model's adversarial robustness but our proposed objective is motivated in the opposite direction. Note that our work is \textbf{not} directly related to TextGANs as well as their applications on dialogue response generation. 

Specifically, the proposed inverse adversarial training assigns higher rewards to generated responses or ground-truth responses if their likelihood decreases more when the dialogue history is perturbed, and penalize the model when it generates responses whose likelihood is almost unchanged given either original or perturbed dialogue history as input. This encourages the model to generate more relevant and informative responses and capture dialogue history better. The proposed IAT algorithm can be used in both supervised and self-supervised fashion (with/without reference response), which can be viewed as a form of reward-augmented maximum likelihood (RAML) method~\cite{norouzi2016reward} that improves the original MLE objective or a rewarding scheme for RL-based text generation algorithms. The inverse adversarial learning framework is also conceptually related to self-adversarial learning~\citep{DBLP:conf/iclr/ZhouGXW020} where the the comparison is made between different checkpoints of the same model to provide reward for RL training of the NLG model.

In addition, we identify a limitation of the widely-used maximum mutual information (MMI) based methods for improving the diversity of dialogue response generation models. This will be discussed in detail in section 2.1 and empirically demonstrated in section 4.2.

We conduct experiments on two dialogue datasets, OpenSubtitiles and DailyDialog, to demonstrate the effectiveness of the proposed approach. Experimental results show IAT helps neural dialogue systems model dialogue history better and generate more diverse and informative responses. 

\section{Related Work}

\subsection{Dull Response Problem}

Neural dialogue models tend to generate generic or dull responses such as \textit{I don’t know} which are not engaging for the users~\cite{sordoni2015neural}. This behavior can be ascribed to the high frequency of generic responses in the training corpus and the over-simplified MLE training objective. How to avoid generic responses and to make the dialogue agent more engaging has been a long-standing problem. Previous work attempts to address this problem with different approaches: 1) \citet{li2015diversity} propose a diversity-promoting objective based on Maximum Mutual Information (MMI). Given source $S$ and target $T$, their approach first generates N-best lists based on $P(T|S)$ and then rerank the list by combining $p(T|S)$ and $\lambda p(S|T)$; 2) \citet{NIPS2018_7452} propose to directly optimize $p(S|T)$ together with $p(T|S)$ with an Adversarial Information Maximization objective; and 3) adversarial learning~\cite{li2017adversarial} and dual adversarial learning~\cite{cui2019dal} based on the intuition that real responses are of high diversity, thus can be distinguished from generated responses which are often dull and generic. There are also other methods using distributional constraints of the target responses~\cite{baheti-etal-2018-generating, csaky-etal-2019-improving} or commonsense knowledge~\citep{DBLP:conf/acl/WuLZZW20}.

While shown to be effective in several datasets, these approaches suffer from several drawbacks. For the first two approaches, while the MMI objective may lead to larger mutual information, it often does not actually result in more informative and engaging responses according to our observations. For example, given a dialog context: ``What have you done with him in the bar last night?'' The top response re-ranked by the MMI objective is ``I have done nothing with him in the bar last night.'', which is non-informative and less natural compared with the response ``Nothing at all.'' generated by a standard seq2seq dialogue model. This is also confirmed in the experiment section. We suspect this phenomenon is caused by the term $p(S|T)$ in the MMI objective. It encourages generating responses that make the last utterance in the dialogue history have a high likelihood given the generated responses. While a truly informative response may yield a high $p(S|T)$, the model can easily find a ``shortcut'' to cheat this objective by simply copying a portion of tokens in the last utterance, which is likely to have high $p(S|T)$ as well as $p(T|S)$. The adversarial learning based dialogue model is notoriously hard to train and may suffer from the problem of mode collapse, which decreases the diversity of generated responses.

In contrast, our proposed IAT approach is based on the intuition that a diverse, relevant, and consistent response should be sensitive to the perturbation in the dialogue history, which is from a different perspective and may be complementary with the aforementioned approaches.

\subsection{Dialogue History Modeling}

Recently, \citet{sankar2019neural} evaluated whether existing neural dialogue systems use dialogue history effectively by perturbing dialogue history and observing the variation of model output. They corrupted the dialogue history with both utterance-level and word-level perturbation and see whether and how much the output perplexity decreases. Their experimental results show that end-to-end neural dialogue systems are generally non-sensitive to the perturbation of dialogue history, suggesting that they may perform poorly in modeling dialogue history. Previous work~\cite{serban2016building,zhao-etal-2017-learning} improves the context modeling ability with modification in model architectures. In contrast, our approach employs a novel training objective to enhance the dialogue history modeling ability, which is orthogonal and may be complementary with them.

\section{Inverse Adversarial Training}

\begin{figure*}[htbp]
\centering
\subfigure{
\centering
\includegraphics[width=0.45\textwidth]{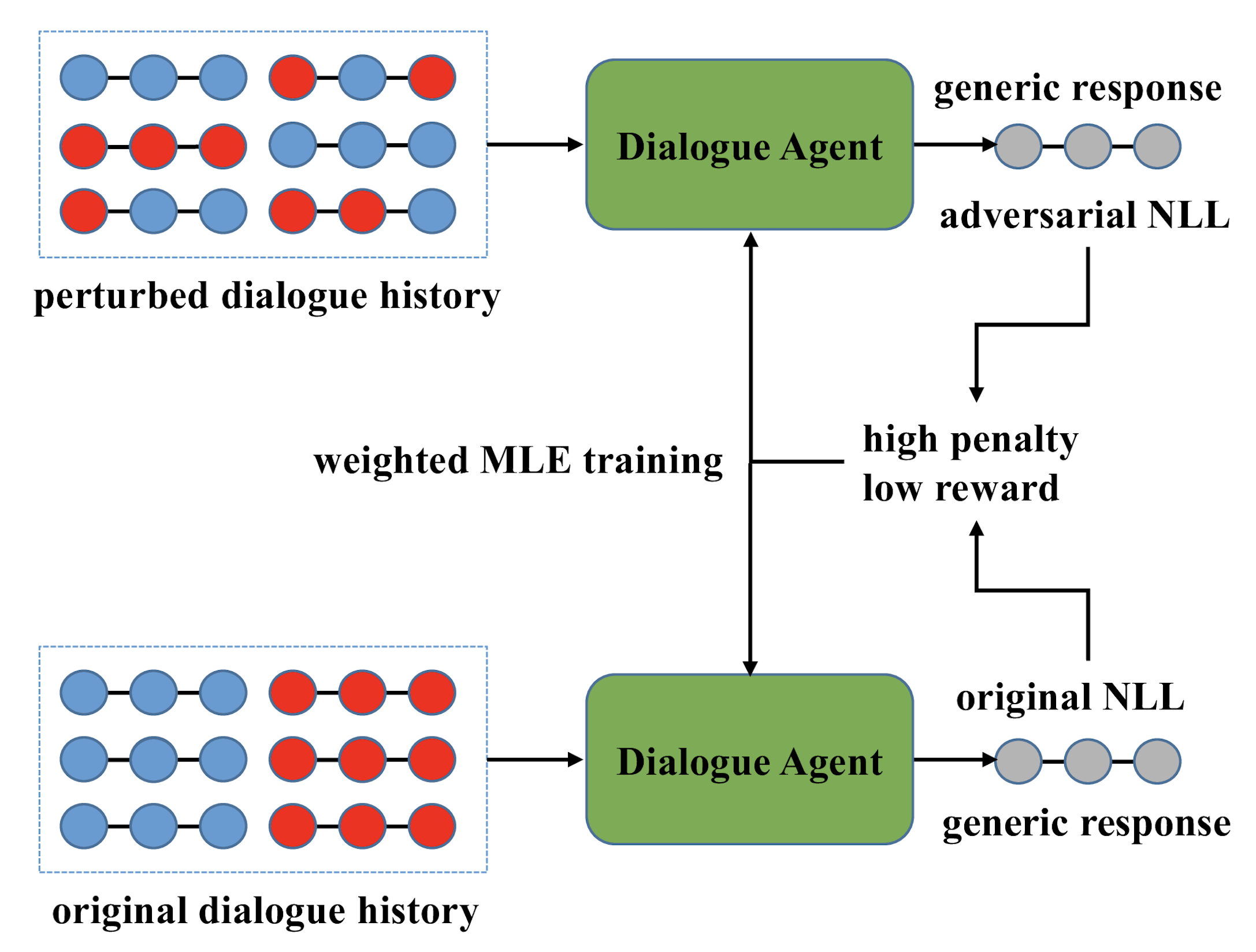}
}%
\subfigure{
\centering
\includegraphics[width=0.45\textwidth]{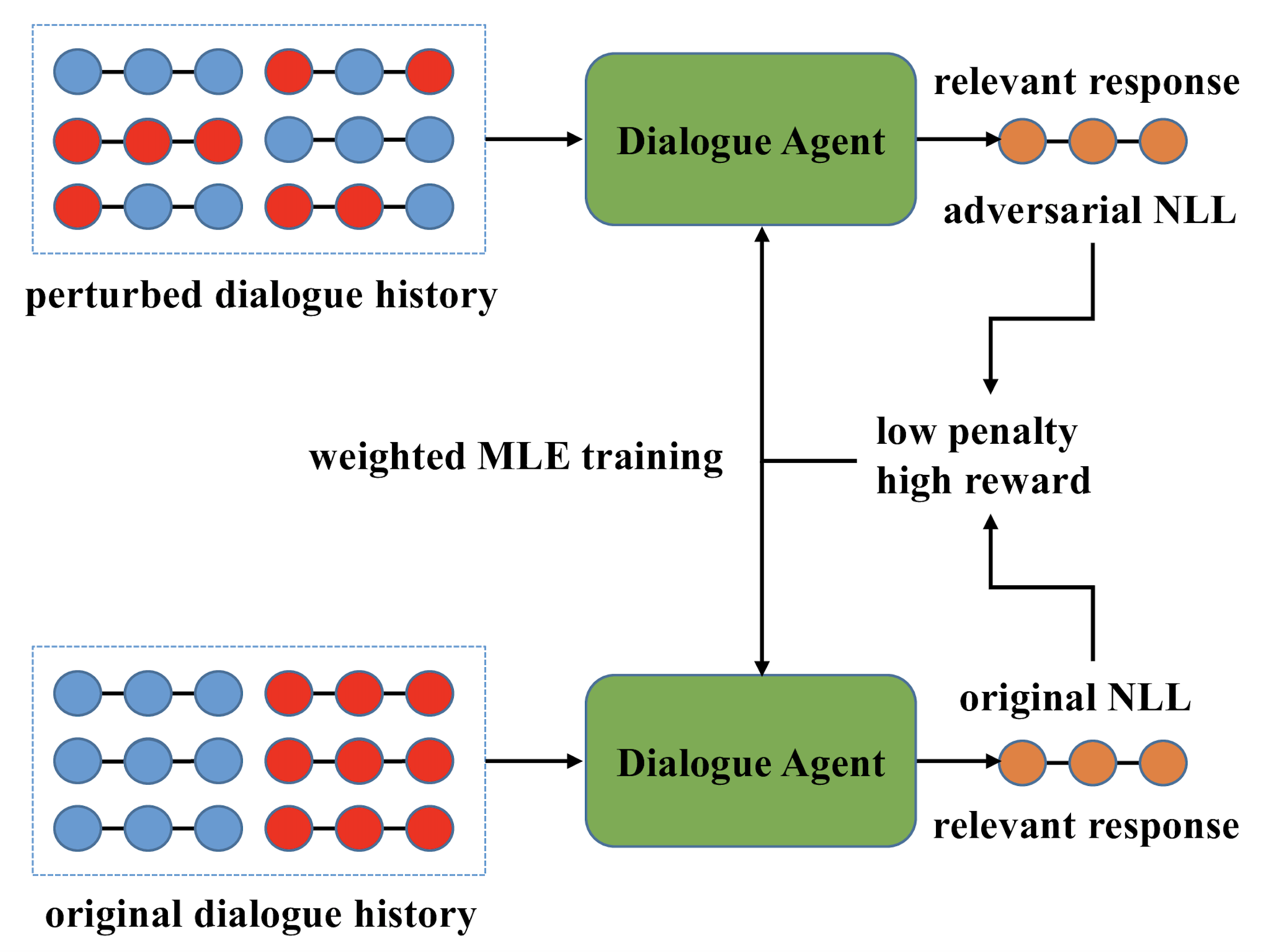}
}%
\centering
\caption{Illustration of IAT. Our algorithm assigns high reward and low penalty when the dialogue model generates relevant and engaging responses given original and perturbed dialogue history respectively. The reward and penalty are respectively decreased and increased when the dialogue model generates dull responses. Note that both dull responses and engaging responses are gold human-written reference responses. They are not labeled in the dataset but automatically detected by the difference of their generation likelihood when given original and perturbed dialogue history. (Best view in color.)}
\end{figure*}

In this section, we describe the proposed inverse adversarial training algorithm in detail. We first describe how we perturb the dialogue history and then formally introduce the inverse adversarial training algorithm.

\subsection{Perturbation Approaches}

Following previous study~\cite{sankar2019neural}, we perturb the dialogue history in both utterance and word level and apply them jointly during training.

\paragraph{Utterance-level Perturbations}

We consider the following operations 1) \textit{Shuf} that shuffles the sequence of utterances in the dialog history, 2) \textit{Rev} that reverses the order of utterances in the history (but maintains word order within each utterance) 3) \textit{Drop} that completely drops certain utterances, 4) \textit{Truncate} that truncates the dialog history to contain only the k most recent utterances where k $\leq$ n, where n is the length of dialog history, and 5) \textit{Repl} that randomly replaces each utterance in the dialogue history by another utterance in the dataset with a probability of 30\%, which resembles the negative sampling~\cite{mikolov2013distributed} approach\footnote{The first four kinds of perturbation is originally proposed in~\cite{sankar2019neural} and the last is proposed in this paper.}.

\paragraph{Word-level perturbations}
We consider similar operations but at the word level within every utterance 1) \textit{word-shuffle} that randomly shuffles the words within an utterance 2) \textit{reverse} that reverses the ordering of words, 3) \textit{word-drop} that drops 30\% of the words uniformly 4) \textit{noun-drop} that drops all nouns, 5) \textit{verb-drop} that drops all verbs, and 6) \textit{word-repl} that replace 30\% of words with a random word in the vocabulary uniformly.

We explain the role of different perturbations and their potential effects briefly. The \textit{Shuf} and \textit{Rev} perturbations change the chronological order of utterances. Inverse adversarial training with these kinds of perturbation may help the model to capture some common-senses about the chronological order of utterances. The \textit{Drop} and \textit{Repl} perturbations may help the model to capture some kinds of casual effects. Finally, the \textit{Truncate} perturbation may help the model capture long-term and multi-turns dialogue history better.

\subsection{Inverse Adversarial Training}

In contrast to the adversarial training objective which maximize the likelihood of generating the same output given perturbed input, the inverse adversarial training objective maximizes the reduction of the likelihood of generating the same output when the input is perturbed, which is opposite to the conventional adversarial training.

A straightforward approach is to maximize the likelihood of generating ground-truth responses given original dialogue history while minimizing this likelihood when given perturbed dialogue history. However, this approach suffers from several problems: First, as a previous study~\cite{sankar2019neural} has shown, neural dialogue models generally capture the perturbation in the dialogue history poorly, which is suggested by the fact that the output embeddings of the encoder are very similar when given original and perturbed input dialogue histories. This results in training the decoder to simultaneously maximize and minimize the likelihood of the same output given very similar input, which is undesirable and makes the training ineffective. The second problem is that this training objective does not capture the variation of likelihood and thus treats relevant and engaging responses equally with dull and generic responses. This is undesirable as we only want to maximize/minimize the likelihood for relevant and engaging responses when conditioning on original/perturbed dialogue history and dull responses should be avoided in both cases.

In this paper, we propose a sequence-level objective which is able to capture the variation of the likelihood of responses given original or perturbed input. This makes it possible to model dialogue history better and avoid generic response problem at the same time. The idea is to evaluate generated sentences based on the variation of the likelihood of responses given original or perturbed dialogue history and use this variation as rewards for training the dialogue model.

Given original dialogue history $X$ and perturbed dialogue history $X'$, the reward $\mathcal{R}(Y|X,X')$ of generating response $Y$, which is a sequence of $n$ tokens $y_{i}, i \in {1,2,...,n}$, is measured by how much $Y$ is more likely to be generated by the dialogue model given $X$ compared with that given $X'$, which is computed by the difference of negative log-likelihood losses (NLL) in two cases, as described below.
\begin{equation}\label{nllo}
 \text{NLL}_{orig} = -\sum_{i = 1}^{n}{\log {P(y_{i}|y_{<i},X)}}
\end{equation}
\begin{equation}\label{nlla}
 \text{NLL}_{adv} = -\sum_{i = 1}^{n}{\log {P(y_{i}|y_{<i},X')}}
\end{equation}
\begin{equation}\label{reward}
 \mathcal{R}(Y|X,X') = \text{NLL}_{adv} - \text{NLL}_{orig}
\end{equation}
Intuitively, the reward $\mathcal{R}$ would be high when the response $Y$ is engaging and relevant to the dialogue history. A generic response should be assigned with a low or even negative reward as it is irrelevant to the dialogue history. The inverse adversarial training objective is to generate responses to maximize its reward. With likelihood ratio~\cite{sutton2000policy}, we can formulate the gradient of the objective function for dialogue response generator $G_{\theta}$ as: 
\begin{equation}
\small
\label{Generator}
\nabla_{\theta}J(\theta) = \sum_{Y}{\sum_{i=1}^{n}\nabla_{\theta}\log G_{\theta}(y_{i}|y_{<i},X) \cdot \mathcal{R}(Y|X,X')}
\end{equation}
The above training objective encourages the dialogue model to generate non-generic responses and model dialogue history better by giving higher rewards when generating good responses based on original dialogue history. 

Similarly, we would also want to penalize the dialogue model when it generates the same response given perturbed dialogue history to explicitly force the dialogue system to effectively model the dialogue history. We propose to model this penalty with a max-margin reward scheme. Given margin $\mathcal{M}$, the penalty $\mathcal{P}(Y|X,X')$ of generating $Y$ is computed by
\begin{equation}\label{penalty}
\small
 \mathcal{P}(Y|X,X') = \min(0, \text{NLL}_{adv} - \text{NLL}_{orig} - \mathcal{M})
\end{equation}
The insight behind equation \ref{penalty} is that when the variation of likelihood of generating $Y$ given $X$ and $X'$ is large enough (i.e. $\text{NLL}_{orig} - \text{NLL}_{adv} - \mathcal{M} > 0$), the model should be considered successfully captured the perturbation in the dialogue history and should not be penalized. In contrast, when the variation is not large enough, we penalize the dialogue agent for generating $Y$ giving $X'$ because a small variation of likelihood implicates: (1) the dialogue agent models dialogue history poorly and (2) the generated responses $Y$ may be irrelevant to the dialogue history $X$ and thus be generic and non-informative. The corresponding gradient can be formulated as:
\begin{equation}
\small
\label{Generator_p}
\nabla_{\theta}J'(\theta) = \sum_{Y}{\sum_{i=1}^{n}\nabla_{\theta}\log G_{\theta}(y_{i}|y_{<i},X') \cdot \mathcal{P}(Y|X,X')}
\end{equation}

The penalty and reward are combined by directly summing up the gradient in Eq (\ref{Generator}) and Eq (\ref{Generator_p}). The proposed inverse adversarial training algorithm can be applied in both supervised fashion where responses $Y$ are ground-truth responses in the dataset and self-supervised fashion where $Y$ is generated by the dialogue model itself. The only difference between the self-supervised and supervised version is whether the reference responses are generated (self-supervised) or ground-truth responses (supervised). The supervised inverse adversarial training can be viewed as a reward function algorithm for RAML~\cite{norouzi2016reward} training that assigns higher rewards for ``good'' training examples that help our model to generate relevant responses and learn to model dialogue history better. The self-supervised inverse adversarial training, in contrast, allows the model to explore freely and train the model with policy gradient~\cite{sutton2000policy}, a reinforcement learning approach.

\section{Experiments}

To validate the effectiveness of the proposed inverse adversarial training algorithm, we conduct experiments in order to answer the following two research questions: 

    \paratitle{(1)} Do inverse adversarial training help neural dialogue systems model dialogue history better?
    
    \paratitle{(2)}  Do inverse adversarial training help neural dialogue models generate more diverse, engaging, and informative dialogue responses?


\subsection{Experimental Settings}

\textbf{Datasets} We employ two datasets in our experiments. The first dataset is the OpenSubtitles corpus~\cite{lison2016opensubtitles2016} which is a large, open-domain dataset containing scripts of movie characters. Following previous work, we consider each turn in the dataset as the target response and the two previous sentences as the dialogue history. We remove the pairs whose response is shorter than 5 words and randomly sample 1,800K, 500K, and 12K dialogue turns for training, validation, and testing, respectively.

We employ the DailyDialog dataset~\cite{li2017DailyDialog} as the second dataset which consists of dialogues that resemble daily conversations across multiple topics. It comprises of 13k dialogues, which is much smaller compared with the OpenSubtitles dataset. However, it has an average of 7.9 turns per dialog, which is more suitable for evaluating whether the proposed approach is able to improve the model's ability of modeling long-term dialogue history.

\textbf{Compared Models} We build dialogue systems with seq2seq~\cite{sutskever2014sequence} models. Following previous work~\cite{li2017adversarial,li2015diversity}, we employ LSTM-based seq2seq model for the OpenSubtitles dataset. For the DailyDialog dataset, we employ the transformer~\cite{vaswani2017attention} model which yields superior results in preliminary experiments while shown to perform poorly in modeling dialogue history~\cite{sankar2019neural}. Specifically, following previous work~\cite{xu2018diversity}, we set the hidden size to 256, embedding size to 128, vocabulary size to 50K, and batch size to 64 for the proposed models and the baselines. We use the Adam optimizer with the initial learning rate 0.1 for model training.

We compare the dialogue model trained with the proposed inverse adversarial learning algorithm with the following baseline methods (all compared models are using the same backbone architecture): 
\begin{itemize}
    \item \textbf{Seq2Seq}: The vanilla seq2seq dialogue model trained with MLE objective.
    \item \textbf{Seq2Seq + MMI}: The dialogue model using mutual information method~\cite{li2015diversity}, which substracts the score of the target sequence $\log p(T | S)$ by its language model score $\log p(T)$ (MMI-anti) or by a backward generation score $\log p(S|T)$ (MMI-bidi) for decoding.
    \item \textbf{Seq2Seq + Adversarial Learning}: A dialogue model trained with adversarial learning objective~\cite{li2017adversarial}. The model is pretrained with MLE objective and then fine-tuned with adversarial learning.
    \item \textbf{Seq2Seq + DS}: A strong baseline using distributional constraints over the generated responses~\cite{baheti-etal-2018-generating}.
    \item \textbf{CVAE}: A dialogue response generation model using conditional VAE~\citep{zhao-etal-2017-learning} to improve the discourse-level diversity of generated responses.
\end{itemize}
    
Our models are pretrained with the MLE objective until the validation perplexity stops decreasing. We then apply the inverse adversarial training algorithm for continual training. During training, reference responses are either generated responses or ground-truth responses in self-supervised and supervised inverse-adversarial training respectively. We combine both supervised and self-supervised inverse adversarial training by alternatively switching between these two objectives for each training iteration. 

\textbf{Evaluation Metrics} We employ different automated evaluation metrics to respectively answer the three research questions introduced at the beginning of this section. To evaluate how well dialogue systems are able to model dialogue history, we adopt the approach proposed by ~\citet{sankar2019neural}, which measures the increases in perplexity when the model is fed with perturbed dialogue history instead of original dialogue history. We report the result in both utterance-level and word-level perturbation.

To evaluate if inverse adversarial learning can effectively reduce the generic response problem, following ~\citet{li2015diversity}, we evaluate the diversity of generated responses by calculating the number of distinct unigrams, bigrams, and trigrams in generated responses. The value is scaled by the total number of generated tokens to avoid favoring long sentences, which are shown as distinct-1, distinct-2, and distinct-3 in Table 2. Lastly, we compare the percentage of
stop-words\footnote{Stopword List from https://www.ranks.
nl/stopwords. We appended punctuations to this list.} of the responses generated by each model (smaller values that are closer to the distribution of human conversations are preferred). We also report the token-level overlap between the generated response and the last utterance in the dialog history to demonstrate the ``shortcut''problem of MMI-based methods decribed in Section 2.1.

As our approach is training in an ``opposite'' direction compared to conventional adversarial training employed to enhance the robustness of trained models, we also conduct experiments to evaluate the robustness of the dialogue response generation models with respect to non label-changing adversarial dialogue history. Similar to the method of evaluating the dialogue history modeling ability, we measure the perplexity changes when the model is given a different but meaning-preserving dialogue history, which is constructed by performing word substitution with a BERT-based lexical substitution method~\cite{zhou2019bert} and paraphrase generation~\cite{sgcp2020} as word-level and utterance-level perturbation respectively on the original dialogue history, as the input.  

In addition, as demonstrated by \citet{liu2016not,DBLP:conf/aaai/Zhou020}, automated metrics are notoriously poor for evaluating dialogue systems.  We thus conduct a human evaluation to better evaluate the effectiveness of the proposed algorithm. For human evaluation, we invite 20 human annotators which are all graduate students with good English proficiency to evaluate the quality of the model. Following~\citet{zhang2018personalizing}, we ask human annotators to interact with compared models for 50 utterances with each compared dialogue system and evaluate the fluency, consistency, and diversity of the model (scored between 1- 5). Fluency measures how likely the generated text is produced by human. Consistency measures how likely the generated text is related to the input dialogue history, which corresponds to the first research question.  Diversity measures how much the generated text provides specific information, rather than “dull” and repeated information, which corresponds to the second research question.

\begin{table}[t!]
\begin{center}
\scalebox{0.95}{
\begin{tabular}{lcc}
\hline\hline
\textbf{Method} & \textbf{DailyDialog} & \textbf{OpenSubtitles} \\ \hline
\bf Seq2Seq &  &  \\
~ - base model & 2.71 (1.18) & 1.94 (0.33) \\
~ - + AL & 2.76 (1.22) & 1.67 (0.41) \\
~ - + DS & 2.79 (1.24) & 1.87 (0.44) \\
~ - + CVAE & 3.25 (1.41) & 2.11 (0.49) \\
~ - + IAT & \bf 3.69 (1.65) & \bf 2.37 (0.42) \\

\hline\hline
\end{tabular}}
\end{center}
\caption{\label{result1} Results on the dialogue history modeling ability of compared models, which is measured by the difference between perplexity of gold responses when receiving original dialogue history and receiving perturbed dialogue history. Mean and standard deviation of 5 runs are reported.}
\end{table}

\begin{table*}[t!]
\begin{center}
\resizebox{\textwidth}{!}{
\begin{tabular}{lccccc|ccccc}
\hline\hline
\multirow{2}{*}{\textbf{Method}} & \multicolumn{5}{c}{\textbf{DailyDialog}} & \multicolumn{5}{c}{\textbf{OpenSubtitles}} \\ 
& \bf Dist-1 & \bf Dist-2 & \bf Dist-3 & \bf overlap & \bf stop-word & \bf Dist-1 & \bf Dist-2 & \bf Dist-3 & \bf overlap  & \bf stop-word \\ \hline
\bf Seq2Seq &  & & & & &  \\ 
~ - base model  & 2.32 & 6.28 & 9.43 & 15.6 & 67.4 & 1.72 & 5.37 & 7.64 & 22.5 & 77.8 \\
~ - + MMI-anti  & \bf 4.15$^{*}$ & \bf 11.27$^{*}$ & \bf 19.61$^{*}$ & 26.7 & 62.4 & 3.45 & 11.35 & 18.12 & 30.1 & 74.2 \\
~ - + MMI-bidi  & 3.52 & 9.29 & 17.43 & 31.5 & 63.1 & \bf 3.52$^{*}$ & \bf 12.11$^{*}$ & \bf 18.56$^{*}$ & 37.8 & 74.7 \\
~ - + AL  & 2.25 & 6.01 & 9.39 & 16.1 & 66.8 & 2.97 & 5.44  &  7.46 & 23.5 & 76.4 \\
~ - + DS  & 3.19 & 7.84 & 11.61 & 18.4 & 61.5 & 3.05 & 6.30  &  11.59 & 21.3 & 71.2 \\
~ - + CVAE  & 3.59 & 9.41 & 12.93 & 17.7 & 61.1 & 3.35 & 10.13 & 17.02 & 22.5 & 71.4 \\
~ - + IAT  & 3.72 & 9.81 & 14.93 & \bf 15.4$^{*}$ & \bf 60.9 & 3.29 & 10.16 & 17.30 & \bf 20.8$^{*}$ & \bf 70.9$^{*}$ \\
\hline\hline
\end{tabular}}
\end{center}
\caption{\label{result1} Results of the diversity of generated responses of compared models. We report the average value of 5 runs on both datasets. $^{*}$ denotes statistically significant with p-value $<$ 0.01.}
\end{table*}

\begin{table}[t!]
\begin{center}
\resizebox{0.9\linewidth}{!}{
\begin{tabular}{lcc}
\hline\hline
\textbf{Method} & \textbf{DailyDialog} & \textbf{OpenSubtitles} \\ \hline
\bf Seq2Seq &  &  \\
~ - base model &  0.75(0.41) & 0.42(0.29) \\
~ - + MMI & - & - \\
~ - + AL & 0.83(0.47) & 0.49(0.34)\\
~ - + DS & 0.78(0.44) & 0.46(0.33)\\
~ - + IAT & 0.77(0.45) & 0.44(0.31)\\
\hline\hline
\end{tabular}}
\end{center}
\caption{\label{resultadv} Results on the adversarial robustness of compared models, which is measured by the difference between perplexity of gold responses when receiving original dialogue history and receiving non-label changing adversarial dialogue history. AL denotes adversarial learning and IAT denotes inverse adversarial training.}
\end{table}

\subsection{Experimental Results}

\textbf{Results on dialogue history modeling} We first present the results on dialogue history modeling ability. The results are shown in Table 1. We can see that the dialogue model trained with the proposed inverse adversarial training algorithm performs significantly better than the compared baselines as the perplexity dramatically increases when the input dialogue history is perturbed. This is not surprising as our approach is the first learning objective which explicitly forces the dialogue system to better model dialogue history. In contrast, the MMI criterion and the adversarial learning objective do not significantly influence the dialogue history modeling ability of dialogue systems. The dialogue model based on CVAE models dialogue history better than other baselines while still under-performs our approach.

\begin{table}[t!]
\begin{center}
\resizebox{1.\linewidth}{!}{
\begin{tabular}{lccc}
\hline\hline
\textbf{Method} & \textbf{Fluency} & \textbf{Consistency} & \textbf{Diversity} \\ \hline
\bf Seq2Seq &  & & \\
~ - base model & 2.83 & 2.69 & 3.05 \\
~ - + MMI-anti & 2.73 & 2.78 & 3.10 \\
~ - + MMI-bidi & 2.80 & 2.82 & 3.02 \\
~ - + AL & 2.77 & 2.69 & 2.91 \\
~ - + DS & 2.85 & 2.88 & 3.12 \\
~ - + CVAE & 2.93 & 2.91 & 3.19 \\
~ - + IAT & \bf 3.02$^{*}$ & \bf 3.05$^{*}$ & \bf 3.34$^{*}$ \\

\hline\hline
\end{tabular}}
\end{center}
\caption{\label{result1} Human evaluation results of compared model on the DailyDialog dataset. The Kappa score between human annotators is 0.73. $^{*}$ denotes statistically significant with p-value $<$ 0.05.}
\end{table}

\textbf{Reults on diversity} The results of the diversity of responses generated by compared models are shown in Table 2. We can see that both the Maximum Mutual Information objective and the proposed inverse adversarial learning succeed in improving the diversity of generated responses. In contrast, the adversarial learning objective hardly improves the diversity, which may be due to the instability of adversarial learning on text generation. While the MMI objective yields slightly larger improvements on distinct n-gram based metrics, their approach is used only for re-ranking during inference, which is orthogonal and may be complementary to the proposed approach. 

\begin{table*}[t!]
\begin{center}
\scalebox{0.85}{
\begin{tabular}{lc}
\hline\hline
\textbf{Source} & how long will it take us to drive to London ? ... \ ok . You haven't seen my company car , have you ? \\ \hline
\bf Seq2Seq &  I don't know about what you are talking about. \\
~ +MMI-anti &  I have seen a car in London.  \\
~ +MMI-bidi &  I haven't seen your company car, have you ? \\
~ +DS &  No, I haven't seen a car. \\
~ +CVAE &  No, I think no. \\
~ +IAT &  No, I don't think we can arrive on time. \\ \hline
\textbf{Source} & Your throat is inflamed . And your tongue is heavily coated . You have all the symptoms of influenza . \\ \hline
\bf Seq2Seq &  I agree with that you have just said. \\
~ +MMI-anti &  I want to have all the influenza. \\
~ +MMI-bidi &  I do not want to have the influenza and the throat. \\
~ +DS &  Oh That's too bad. \\
~ +CVAE &  That's a sad thing. \\
~ +IAT &  oh what should I do for my tougue? \\
\textbf{Source} & Good evening , madam . Can I help you ? Yes . I bought this sheet here yesterday . \\ \hline
\bf Seq2Seq &  I want to buy a sheet too. \\
~ +MMI-anti &  What is the color of the shirt you bought yesterday? \\
~ +MMI-bidi &  What is wrong with this sheet bought here yesterday? \\
~ +DS &  What problem about the sheet? \\
~ +DS &  What do you think of the sheet? \\
~ +IAT &  Anything wrong with the sheet? \\ 
\hline\hline
\end{tabular}}
\end{center}
\caption{\label{result1} Case study of dialogue responses generated by different compared models.}
\end{table*}

In addition, as described in section 2.1, the MMI objective may favor non-engaging responses that simply repeats the last utterance in the dialogue history. This is empirically demonstrated by their high overlap with the last utterance in the dialog history, as measured by the ``overlap'' metric. In contrast, our approach does not suffer from this problem and also generate fewer stop-words compared to the MMI-based methods. In addition, our approach also outperforms the strong baselines including that using distributional constraint and CVAE, demonstrating its effectiveness in improving the diversity of generated responses. 

\textbf{Results on adversarial robustness} We also conduct experiments to test the robustness of the dialogue model trained with the proposed inverse adversarial training objective. The results are shown in Table 3. We see that the increase in the perplexity of ground-truth responses under our model is roughly the same with the baseline transformer model and the other compared models. This suggests that our proposed IAT objective does not harm the adversarial robustness.


\textbf{Human evaluation} We conduct a human evaluation of compared models on the DailyDialog dataset. The results are shown in Table 4. We can see that the proposed inverse adversarial training objective substantially improves the consistency of the dialogue model over all compared baselines, which confirms its ability to train dialogue agents to model dialogue history better. As for the diversity of generated responses, we find that human annotators do not prefer the responses selected by the MMI objective over that generated by the baseline model with a large margin. We find that this is mainly because the MMI objective prefers repeating tokens which appear in the last utterance and human annotators find it non-informativeness. In contrast, our approach yields even larger improvements in the diversity of the generated responses. We do not find the adversarial learning method improves the diversity of dialogue models, which may be due to the problem of mode collapse in adversarial learning. The over-all fluency of compared models is roughly the same, which may be because they are all trained or pretrained with MLE objective.

\begin{table}[t!]
\begin{center}
\resizebox{1.\linewidth}{!}{
\begin{tabular}{lccc}
\hline\hline
\textbf{Method} & \textbf{Fluency} & \textbf{Consistency} & \textbf{Diversity} \\ \hline
\bf Ours &  2.86 & \bf 3.02 & \bf 3.36 \\
~ - w/o supervised &  2.68 &  2.91 & 3.34 \\
~ - w/o self-supervised &  \bf 2.91 &  2.98 & 3.31 \\ 
~ - w/o reward &  2.74 &  2.88 &  3.12 \\
~ - w/o penalty &  2.88 &  2.82 &  3.26 \\
~ - w/o utter\_pertub &  2.82 &  2.71 & 3.23 \\
~ - w/o token\_pertub &  2.85 &  2.84 &  3.28 \\
\hline\hline
\end{tabular}}
\end{center}
\caption{Ablation study results of compared model on the DailyDialog dataset. AL denotes adversarial learning and IAT denotes inverse adversarial training.}
\label{result1} 
\end{table}

\subsection{Qualitative Analysis}

To better compare and analyze the inverse adversarial training objective, we conduct a qualitative analysis of dialogue responses generated by different compared models. The samples are presented in Table 5. We can see that the vanilla transformer-based dialogue response generation model tends to generate irrelevant and generic responses. Applying the MMI objective for re-ranking successfully avoids those generic responses. However, it leads to another kind of non-informative response that repeats the majority of tokens in the latest utterance, which is also quite unnatural. In contrast, dialogue models trained with the proposed inverse adversarial training objective tend to generate more diverse responses which are also more relevant to the dialogue history. 

\subsection{Ablation Study}

To better understand the relative importance of different components in the proposed inverse adversarial training objective, we conduct an ablation study with human evaluation to compare different model variants against the full model. The results are shown in Table 6. We can find that both supervised-only and self-supervised-only variant of the proposed inverse adversarial training algorithm can improve the consistency and the diversity of dialogue models. However, self-supervised inverse adversarial training seems to sacrifice the fluency of generated responses for better diversity and consistency as the model trained without the self-supervised objective are considered to be more fluent by human annotators. The usefulness of the reward and the penalty objectives is also demonstrated by human evaluation. Concretely, we find that the reward described in Eq.(3) contributes more to the diversity of generated responses. This may be because it assigns high rewards for relevant and specific responses and negative rewards for generic responses. In contrast, the penalty in Eq.(5) helps the dialogue system model dialogue history better and leads to more consistent responses by punishing the dialogue model when generating the same responses given perturbed dialogue history. As for different perturbation approaches, we find that both utterance-level and token-level contributes to the performance improvements. Also, we find that utterance-level perturbation may be more effective for improving the consistency of generated responses. We suspect this may be because the ability of the dialogue model to distinguish utterance-level perturbation is more important for better dialogue history modeling.

\section{Conclusion}

In this work, we introduce inverse adversarial training (IAT) algorithm that is able to simultaneously reduce the dull response problem and help neural dialogue systems model dialogue history better. IAT measures the relevance and consistency of responses by the difference of their likelihood conditioning on either original and perturbed dialogue history. In this way, it is able to prevent the dialogue system from preferring generic responses, even they are often of high frequency in the training corpora. Our method also encourages the dialogue agent to model dialogue history better by penalizing the model when generating the same responses given perturbed dialogue history. Experimental results on two benchmark datasets show that the proposed inverse adversarial training algorithm helps dialogue models capture dialogue history better and generate more diverse and consistent responses. We also identify a limitation of the widely-used MMI based methods for improving the diversity of dialogue response generation models and empirically demonstrate the existence of this problem through our experimetns.

\section*{Boarder Impact}

This work does not involve collection and release of data, nor inference of information or judgments about individuals. However, dialogue systems may have a social impact and we believe that making dialogue agent able to generate more meaningful and consistent responses are beneficial. We also agree that general control on the bias or unfairness of neural dialogue agents is important. We believe this can be done from both the perspective of data collection and training algorithms. We believe our proposed training algorithm will likely not contribute to any ethical concern of chat robots.

\section*{Acknowledgments}
We thank the anonymous reviewers for their valuable comments.



\bibliography{acl21}
\bibliographystyle{acl21}

\end{document}